\documentclass{article} 
\usepackage{iclr2026_conference,times}

\usepackage{amsmath,amsfonts,bm}

\def\eqref#1{equation~\ref{#1}}

\def\1{\bm{1}}

\DeclareMathAlphabet{\mathsfit}{\encodingdefault}{\sfdefault}{m}{sl}
\SetMathAlphabet{\mathsfit}{bold}{\encodingdefault}{\sfdefault}{bx}{n}

\usepackage[utf8]{inputenc} 
\usepackage[T1]{fontenc}    
\usepackage{amsfonts}       
\usepackage{nicefrac}       
\usepackage{microtype}      
\usepackage{makecell}

\usepackage{hyperref}
\usepackage{url}
\usepackage{graphicx}
\usepackage{multirow}
\usepackage[table,xcdraw]{xcolor}
\usepackage{amsmath}
\usepackage{booktabs}
\usepackage[capitalize,noabbrev]{cleveref}
\usepackage{colortbl}
\usepackage{pifont}
\usepackage[normalem]{ulem}
\usepackage{wrapfig}
\usepackage[many]{tcolorbox}
\usepackage{listings}
\usepackage{algorithm}
\usepackage{algorithmic}
\usepackage{caption}

\title{EmoCaliber: Advancing Reliable Visual Emotion Comprehension via Confidence Verbalization and Calibration}

\author{
    \vspace{-0.8cm}\\
    \textbf{Daiqing Wu}$^{1,3}$\quad
    \textbf{Dongbao Yang}$^{1}$\quad
    \textbf{Can Ma}$^{1}$\quad
    \textbf{Yu Zhou}$^{2}$\\
    \vspace{-0.3cm}\\
    \footnotesize{$^1$IIE, Chinese Academy of Sciences}\quad
    \footnotesize{$^2$Nankai University}\quad
    \footnotesize{$^3$University of Chinese Academy of Sciences}\vspace{0.1cm}\\
    \texttt{wudaiqing@iie.ac.cn}\\\\
    \vspace{-1cm}
}

\iclrfinalcopy 
\begin{document}

\maketitle

\begin{abstract}
Visual Emotion Comprehension (VEC) aims to infer sentiment polarities or emotion categories from affective cues embedded in images. In recent years, Multimodal Large Language Models (MLLMs) have established a popular paradigm in VEC, leveraging their generalizability to unify VEC tasks defined under diverse emotion taxonomies. While this paradigm achieves notable success, it typically formulates VEC as a deterministic task, requiring the model to output a single, definitive emotion label for each image. Such a formulation insufficiently accounts for the inherent subjectivity of emotion perception, overlooking alternative interpretations that may be equally plausible to different viewers. To address this limitation, we propose equipping MLLMs with capabilities to verbalize their confidence in emotion predictions. This additional signal provides users with an estimate of both the plausibility of alternative interpretations and the MLLMs' self-assessed competence, thereby enhancing reliability in practice. Building on this insight, we introduce a three-stage training framework that progressively endows with structured reasoning, teaches to verbalize confidence, and calibrates confidence expression, culminating in EmoCaliber, a confidence-aware MLLM for VEC. Through fair and comprehensive evaluations on the unified benchmark VECBench, EmoCaliber demonstrates overall superiority against existing methods in both emotion prediction and confidence estimation. These results validate the effectiveness of our approach and mark a feasible step toward more reliable VEC systems. Project page: \href{https://github.com/wdqqdw/EmoCaliber}{https://github.com/wdqqdw/EmoCaliber}.
\end{abstract}

\section{Introduction}

Perceiving and interpreting emotional signals from visual stimuli is a critical ability that enables humans to communicate empathetically, facilitating smoother information exchange and enhancing interpersonal trust \citep{motiv1}. Consequently, this ability is indispensable to emotionally intelligent agents, whose applications span mental health support \citep{app1}, affective companionship \citep{app2}, and embodied service robotics \citep{app3}. Accompanying this need, Visual Emotion Comprehension (VEC) has attracted rapidly growing attention in recent years \citep{tpami2021aica}. This task requires models to analyze sentiment polarities or recognize the underlying emotions of given images, serving as a cornerstone for developing and evaluating visual emotional intelligence in agents.

Prior research on VEC primarily focuses on small-scale, task-specific models, which typically rely on fine-tuning to recognize a predefined set of emotion categories and learn the corresponding image–emotion mapping. Although these models have achieved notable success, their capabilities are tightly bound to the training objectives, resulting in limited generalizability even to closely related settings \citep{domain2020}. In contrast, the recent emergence of Multimodal Large Language Models (MLLMs) offers a potentially more flexible solution for VEC. By integrating multimodal encoders with powerful Large Language Models (LLMs) \citep{llava2023}, MLLMs inherit strong instruction-following abilities that support the execution of a diverse range of vision tasks through a unified interface \citep{mllm_survey2024}. This makes them intrinsically more adaptable to the complex and continually evolving scenarios in real-world applications.

\begin{figure}[t]
    \centering
    \vskip -0.1in
    \includegraphics[width=1\linewidth]{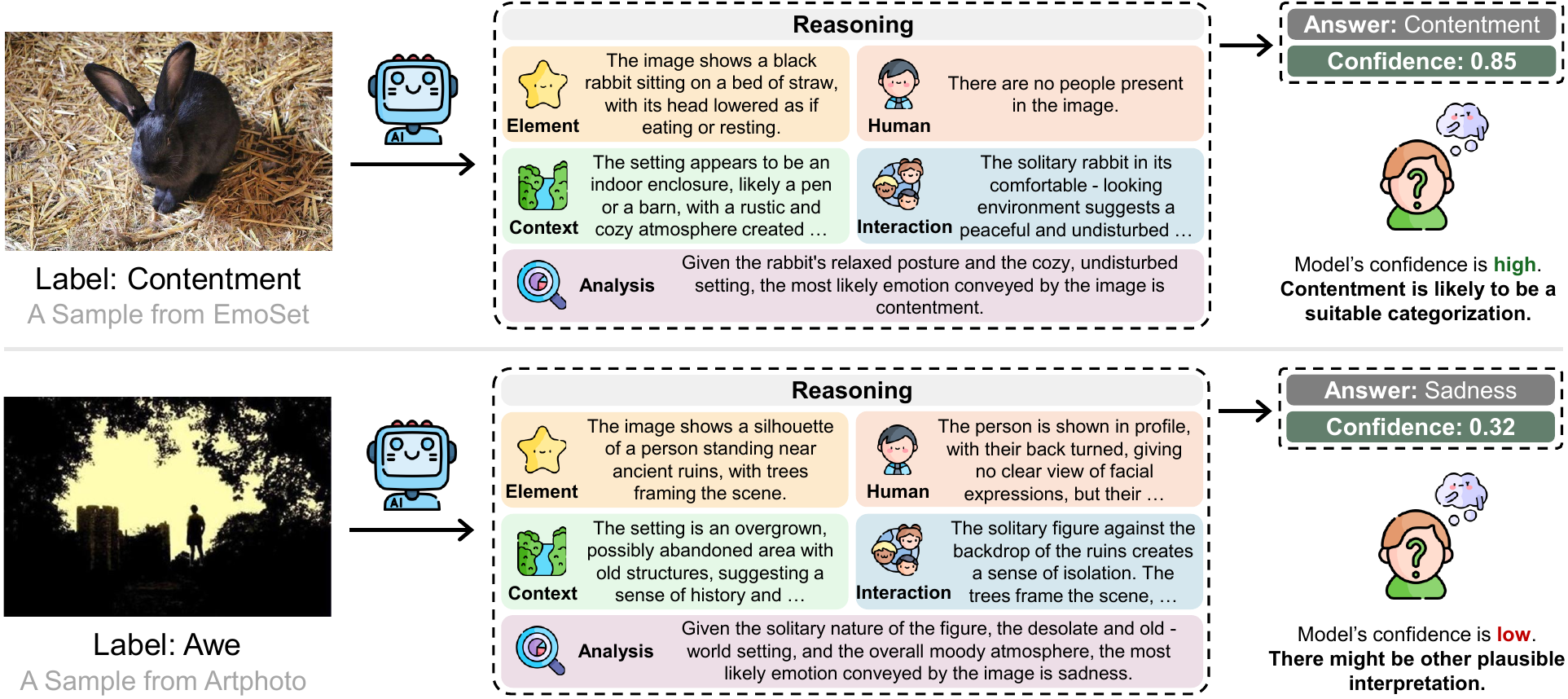}
    \caption{In addition to a structured Chain-of-Thought (CoT) and the derived answer, EmoCaliber also produces a self-evaluated confidence level. It enables users to adopt the output selectively, significantly enhancing the reliability of the VEC system.}
    \vskip -0.1in
    \label{fig:m1}
\end{figure}

Recognizing these advantages, recent research has gradually shifted its focus toward optimizing VEC capabilities within MLLMs. EmoVIT \citep{emovit2024} represents a seminal effort in this direction by introducing the first emotion-centric dataset for visual instruction tuning. After it, a growing body of research has explored complementary perspectives, including training-free adaptations \citep{emnlp2024visual_prompt, icml2025catch}, tuning-based optimization \citep{emollm2024, emo-llama2024}, and benchmarking specific properties \citep{eemobench, customizing}. Despite the current prosperity, these approaches mainly require MLLMs to produce deterministic responses for each query, framing VEC as a definitive task. This formulation overlooks the inherent uncertainties arising from the perceptual subjectivity of emotion \citep{subjectivity}, where a single image may evoke distinct emotional responses from different viewers \citep{personal2016mm}, or even from the same viewer under varying circumstances. Such oversight significantly compromises their reliability in downstream applications, many of which are highly sensitive due to their human-centric nature. 

In this manuscript, we introduce a feasible solution to this issue by teaching MLLMs to explicitly verbalize their confidence levels after generating responses. As shown in \cref{fig:m1}, the model is enabled to assign high confidence for responses supported by clear emotional cues and self-evaluated as accurate, while assigning low confidence to responses involving emotionally ambiguous or beyond its competence. This confidence dimension serves as an auxiliary indicator alongside reasoning and categorization, allowing users to selectively trust or utilize model outputs and thereby mitigating the unreliability caused by the previous definitive formulation.

\begin{figure}[t]
    \centering
    \includegraphics[width=0.95\linewidth]{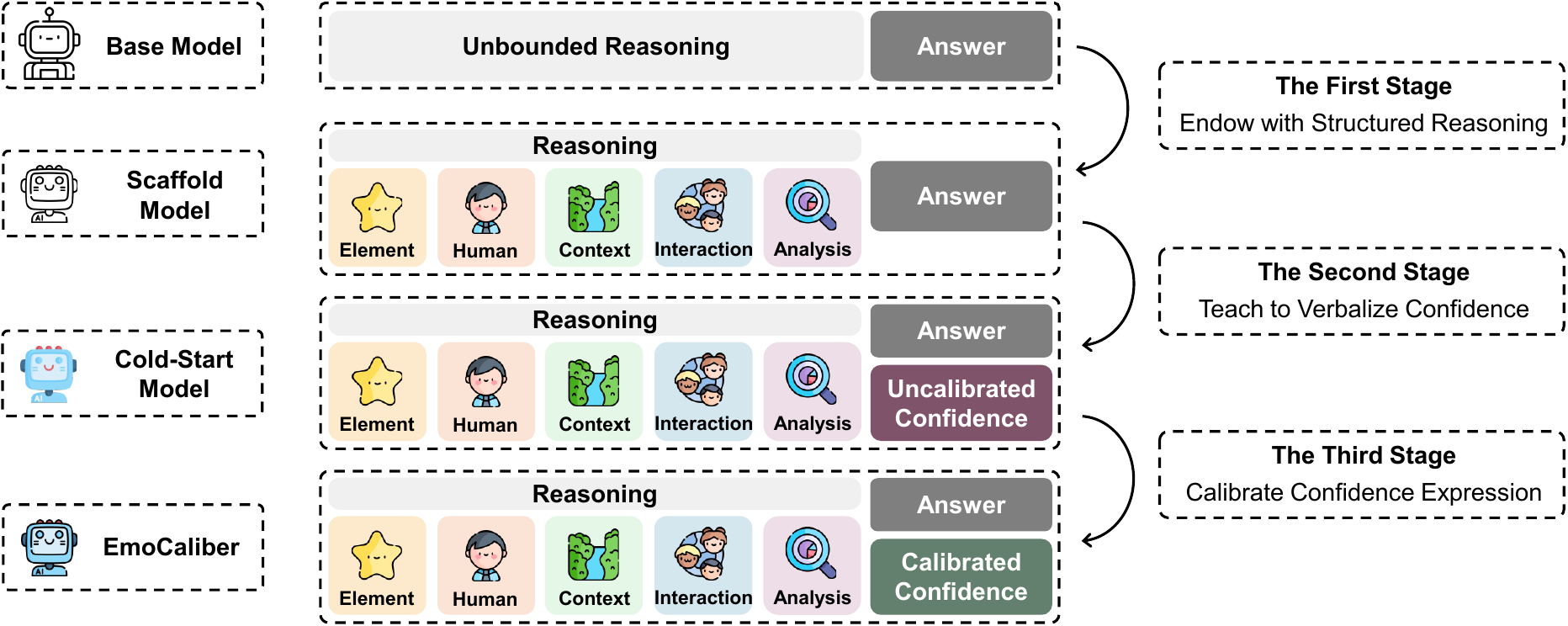}
    \caption{Illustration of the model’s evolution across the three training stages. Through these stages, the model is successively endowed with structured reasoning, taught to verbalize confidence, and finally calibrated to express confidence accurately.}
    \vskip -0.1in
    \label{fig:m2}
\end{figure}

To instantiate this capability, we devise a three-stage training framework, as illustrated in \cref{fig:m2}. In the first stage, we endow the model with structured reasoning capabilities via Supervised Fine-Tuning (SFT). Inspired by the human affective cognition, we formalize the process from visual perception to affective categorization into five structured steps: (1) identifying prominent visual elements in the image; (2) providing detailed descriptions of human subjects, if present; (3) describing contextual elements beyond the subjects; (4) discussing how these elements interact; and (5) deriving an emotional conclusion based on the preceding observations. By structuring the reasoning process, we encourage the model to integrate comprehensive visual evidence before producing well-grounded predictions. We denote the obtained model as the scaffold model.

In the second stage, we further equip the model with the ability to verbalize confidence after the answer. Using a set of data unseen in stage one, we first task the scaffold model to generate structured reasoning. Subsequently, we append each reasoning path with a confidence value, quantified by the deviation between the predicted emotion and the ground-truth label. We then perform SFT of the scaffold model on this augmented data to teach it to verbalize confidence, yielding an advanced version that we name the cold-start model.

In the third stage, we adopt Reinforcement Learning (RL) to calibrate the model's confidence expression, while concurrently incentivizing more precise affective reasoning. Specifically, we design a three-component reward function to jointly optimize response format, emotion prediction, and confidence verbalization. By applying Group Relative Policy Optimization (GRPO) \citep{grpo} on another held-out dataset, we ultimately obtain \textbf{EmoCaliber}, an MLLM proficient for VEC that not only achieves accurate emotion predictions via structured reasoning, but also provides calibrated self-assessment through confidence verbalization.

\begin{figure}[t]
    \centering
    \includegraphics[width=0.85\linewidth]{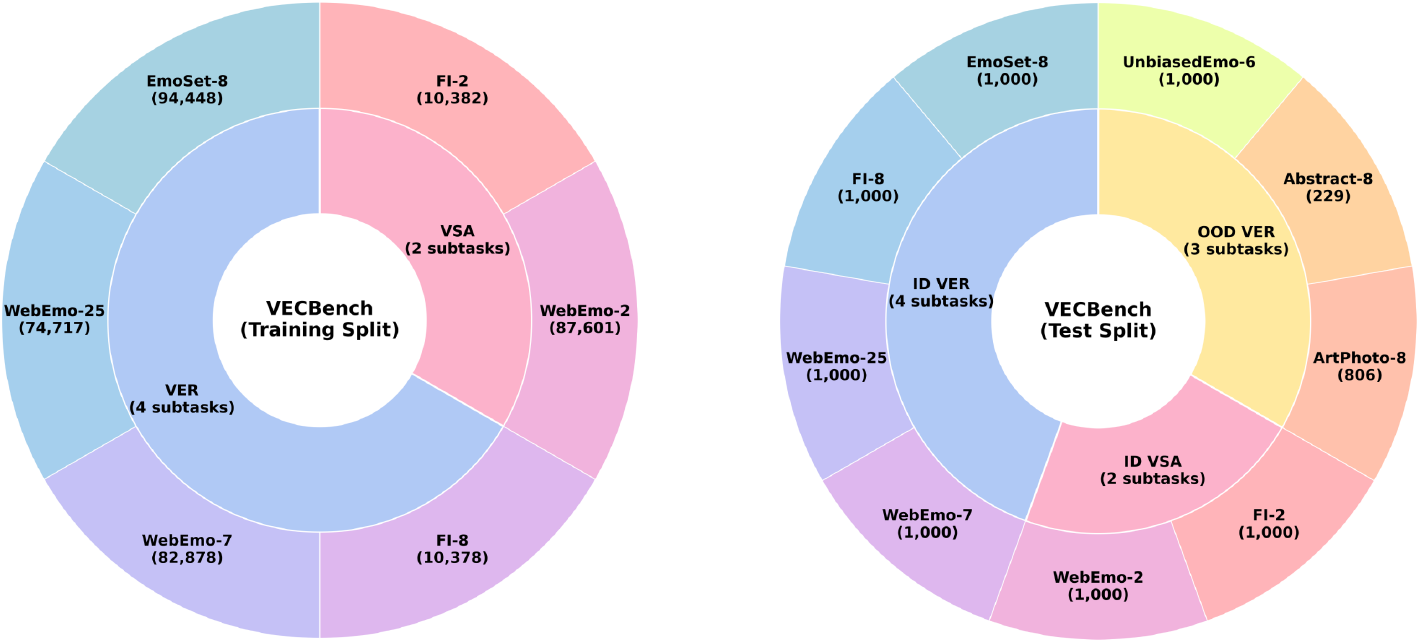}
    \caption{Task composition of VECBench. The training split comprises Visual Sentiment Analysis (VSA) and Visual Emotion Recognition (VER) tasks, with each subtask denoted as ``source-granularity (\#sample)''. In addition to retaining corresponding subtasks for in-domain (ID) evaluation, the test split also includes out-of-domain (OOD) VER tasks to verify generalization ability.}
    \vskip -0.1in
    \label{fig:m3}
\end{figure}

To ensure a systematic and comprehensive evaluation, we group six widely recognized VEC datasets into \textbf{VECBench}. Based on their taxonomies and annotations, we divide them into non-overlapping training and test splits. The training split comprises data from FI \citep{fi2016}, WebEmo \citep{webemo2018}, and EmoSet \citep{emoset2023}, covering Visual Sentiment Analysis (VSA) and Visual Emotion Recognition (VER) tasks. Meanwhile, the test split not only retains in-domain data for evaluation but also incorporates samples from Abstract \citep{abstract2010}, Artphoto \citep{abstract2010}, and UnbiasedEmo \citep{webemo2018} for out-of-domain generalization evaluation. The overall task composition is illustrated in \cref{fig:m3}. By constructing VECBench, we aim to provide a fair and reproducible baseline to facilitate future research on MLLMs for VEC. In summary, our contributions are threefold:

\begin{itemize}
    \item We introduce a practical solution to mitigate unreliability in the current definitive formulation of VEC by enabling MLLMs to verbalize their self-estimated confidence for each response.
    \item To instantiate this capability, we devise a three-stage training framework, which gradually endows structured reasoning, teaches confidence verbalization, and calibrates confidence expression. This ultimately leads to EmoCaliber, a model capable of performing VEC reliably and accurately.
    \item Through a comprehensive evaluation on VECBench, we demonstrate that EmoCaliber delivers not only significant advantages in accurately self-evaluating confidence, but also consistent superiority in both in-domain and out-of-domain VEC performance.
\end{itemize}

The remainder of this paper is organized as follows. \cref{sec:related} reviews prior work closely related to this study. \cref{sec:training} describes the proposed training framework in detail. \cref{sec:exp} presents extensive experimental results, including comparisons with state-of-the-art MLLMs and ablation studies. Finally, \cref{sec:conclusion} concludes the paper.


\section{Related Works}
\label{sec:related}

\subsection{Visual Emotion Comprehension}

\subsubsection{Small-Scale Specialized VEC Models}
As a long-standing research field, the development of VEC has largely mirrored the broader evolution of Artificial Intelligence. Early VEC studies primarily rely on hand-crafted features, exploring cues from low-level textures to mid-level aesthetic attributes and high-level semantic representations \citep{handcraft2014}. With the rise of pretrained models, researchers naturally explore their application to VEC, where the remarkable performance improvements validate the clear benefits of general-purpose knowledge for perceiving emotions \citep{pretrained2017, pr2023msa}. Subsequent studies further advance this direction, typically by introducing human priors through elaborating modules to better capture spatial clues and leverage pretrained knowledge \citep{pr2018face, pr2026code}. In recent years, the proliferation of vision–language pretraining has likewise stimulated progress in VEC, with researchers enhancing visual models through distilling emotional cues embedded in language models \citep{pacl2024, pr2025vlalignment}. Despite achieving notable progress, their training paradigm imposes an implicit ceiling on generalization. Models trained on one emotion taxonomy cannot be directly applied to another, and even within the same taxonomy, performance degrades substantially when evaluated on images from different domains \citep{domain2020}. This limitation constrains their applicability in scenarios that demand greater flexibility.

\subsubsection{VEC Systems based on Multimodal Large Language Models}
The recent emergence of MLLMs offers a potential solution to this dilemma. At the cost of some computational efficiency, VEC systems based on MLLMs achieve unprecedented flexibility in comprehending diverse task formulations, while maintaining a reasonable level of expertise supported by web-scale pretraining \citep{icl2025icml}. Recognizing these advantages, extensive research has emerged to investigate MLLM-based VEC. A series of works focuses on tuning-based optimization, exploring how to inject emotional insights or sharpen emotion perception. For instance, EmoVIT \citep{emovit2024} synthesizes the first emotion-centric instruction–tuning dataset using an advanced proprietary LLM; EmoLLM \citep{emollm2024} further scales up the size of fine-tuning data and designs modules to capture clues from content and relational viewpoints; EMO-LLaMA \citep{emo-llama2024} targets facial expression recognition, adopting a similar paradigm while augmenting it with expert feature extraction and contextual modeling. Another line of research avoids modifying MLLM parameters and instead leverages hierarchical prompting or MLLMs' intermediate states to mirror human affective reasoning. SoV \citep{emnlp2024visual_prompt} breaks down emotion perception into subtasks and iteratively guides MLLMs through a systematically organized set of prompts; SEPM \citep{icml2025catch} employs a two-stage inference pipeline that first uncovers correlations among visual emotions and the MLLM’s internal confidence, then applies tailored strategies to produce appropriate prompts.

\subsubsection{Handling Subjectivity in Emotion Perception}
In MLLM-based VEC research, relatively limited effort has been devoted to handling the inherent subjectivity of emotion perception. This challenge stems from the fact that the same visual stimulus may evoke different emotional responses, depending on factors such as the viewer’s personality, cultural background, or current circumstances \citep{pieee-survey2023}. While mainstream VEC datasets use majority voting to reflect a degree of perceptual consensus, cases where plausible interpretations are overlooked remain inevitable. As illustrated in the lower sample in \cref{fig:m2}, the model's explanation for sadness may resonate with some individuals while still contradicting the annotated label of awe.

A recent notable attempt to mitigate this issue during evaluation is the MVEI benchmark \citep{customizing}, which reformulates the 
classification tasks originally designed for specialized models into an emotion statement judgement task tailored for MLLMs. This reformulation preserves evaluation depth through the expressive flexibility of statements, while reducing the impact of subjectivity by discretizing the answer space. This challenge also receives focus from an adjacent research field, multimodal sentiment analysis. OV-MER \citep{ov-mer} directly annotates videos with multiple open-vocabulary emotion labels via a human–LLM collaboration strategy, thereby capturing the potential diversity induced by subjective perception. 

While inspiring, these approaches rely on training or evaluation protocols that differ from existing mature frameworks, which may limit the direct reuse of well-established baselines and subtask designs. In light of this, we propose teaching the model to verbalize its confidence level as an alternative perspective for handling subjectivity. This approach not only provides users with a quantitative sense of potential subjective interpretations but also aligns with established frameworks.

\subsection{Confidence Verbalization and Calibration}
Since LLMs and MLLMs frequently generate hallucinations or fail to express uncertainty when faced with unfamiliar problems, eliciting the confidence of their responses has become a crucial means to enhance their practical reliability. To achieve this, early attempts \citep{ask2023emnlp} rely on prompt engineering or consistency across sampled answers, which are limited by modest calibration effectiveness or significantly increased latency. Subsequent tuning-based methods introduce group-based \citep{group_uncertainty} or binary \citep{t-tuning} confidence estimation, yet still fall short of capturing sample-wise, fine-grained uncertainty. The recent popularity of RL has further inspired new directions. For instance, SaySelf \citep{sayself} establishes a solid baseline by performing confidence calibration via RL on top of SFT; and RLCR \citep{rlcr} demonstrates the benefit of directly incorporating a Brier score \citep{brier} into the reward function. In VEC tasks, beyond their established role in enhancing reliability, confidence verbalization offers an additional intuitive advantage in reflecting the perceptual subjectivity of emotion. This motivates us to incorporate such a capability into the MLLM-based VEC system.

\section{Training Framework}
\label{sec:training}

Building upon these insights, this section introduces a three-stage training framework, with each stage detailed in \cref{subsec:3_1}, \cref{subsec:3_2}, and \cref{subsec:3_3}.

\subsection{Stage 1: Endowing with Structured Reasoning}
\label{subsec:3_1}

Inferring emotions from visual cues is inherently complex, requiring the integration of salient elements, background contexts, and their interactions. This poses a substantial challenge for generic MLLMs, whose Chain-of-Thoughts (CoTs) are typically unconstrained, leading to diffuse trajectories and neglect of critical clues. To enable structured reasoning with hierarchical clue modeling, we first perform SFT on curated CoT-annotated data.

\begin{figure}[t]
    \centering
    \includegraphics[width=1\linewidth]{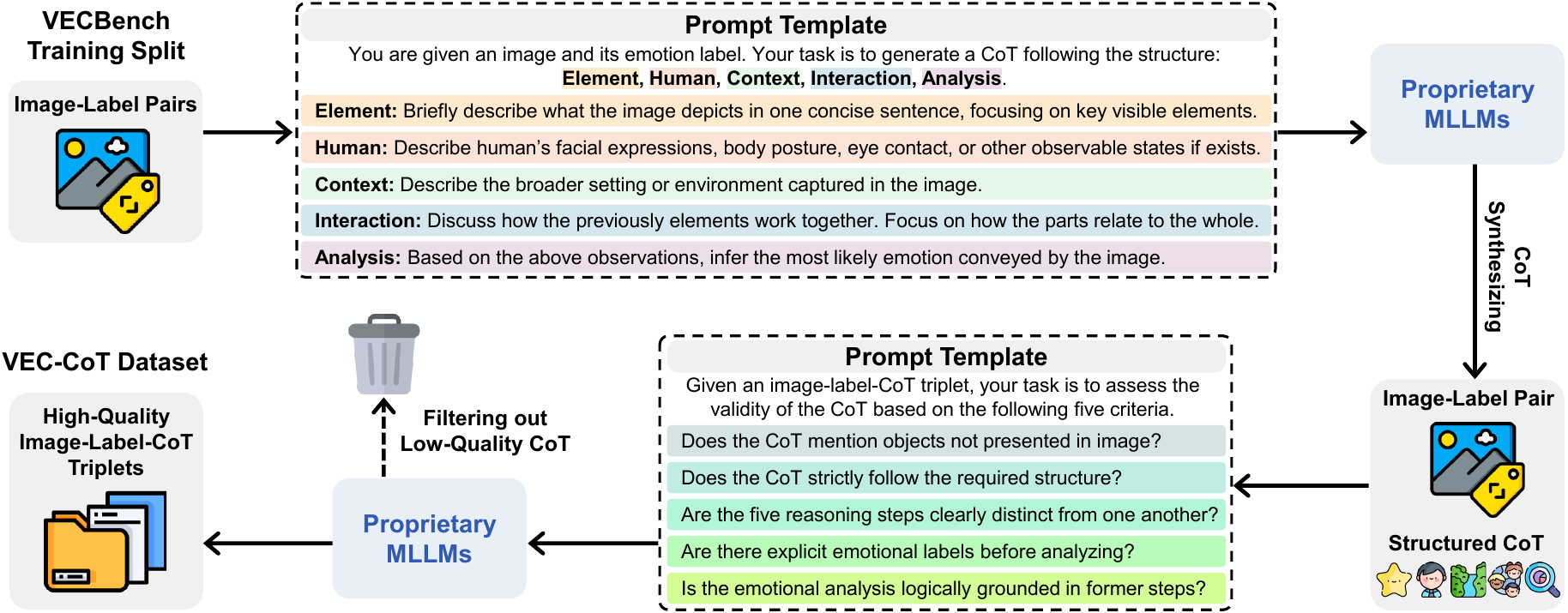}
    \caption{Construction pipeline of the VEC-CoT dataset from the training split of VECBench. Image–label pairs are first templatized and fed into proprietary MLLMs to synthesize structured CoTs, which are then subjected to strict quality evaluation. Finally, image–text pairs with high-quality CoTs are retained and grouped into the VEC-CoT dataset.}
    \label{fig:m4}
\end{figure}

Due to the lack of suitable data, we begin by constructing the VEC-CoT dataset, with its overall pipeline illustrated in \cref{fig:m4}. Inspired by human affective cognition \citep{emotion_cognition}, we decompose VEC into five sequential steps: \textit{1) Salient Element Identification}. Detecting the primary visual elements that initially attract attention, which form the raw evidence for subsequent appraisal; \textit{2) Human Subject Description}. Detailing any present human figures, as they often convey the most explicit and socially anchored emotional signals; \textit{3) Contextual Expansion}. Extending the observation to broader environmental features, which modulate how identical elements may evoke differing interpretations under varying circumstances; \textit{4) Interaction Modeling}. Examining the relationships and causal dynamics among the observed elements, integrating how actions, objects, and settings reinforce or contradict each other; \textit{5) Comprehensive Analysis}. Consolidating the entire evidence chain to derive the final emotional conclusion, thereby bridging low-level visual perception with high-level affective categorization.

\begin{table}[t]
\vskip -0.02in
\caption{Statistics of the VEC-CoT dataset used across training stages. The dataset is split by 6:3:1.}
\vskip -0.1in
\label{m:tab1}
\centering
\resizebox{0.85\linewidth}{!}{
\renewcommand{\arraystretch}{1.1}
\begin{tabular}{cccccccc}
\toprule
\multirow{2}{*}{\bf VEC-CoT} & \multicolumn{4}{c}{VER}                 & \multicolumn{2}{c}{VSA} & \multicolumn{1}{c}{\multirow{2}{*}{Total}} \\
\cmidrule(l){2-5} \cmidrule(l){6-7}
 & EmoSet-6 & WebEmo-25 & WebEmo-7 & FI-8  & WebEmo-2     & FI-2     &  \\
\midrule
1st Stage                  & 30,022   & 12,371    & 15,552   & 3,516 & 21,298       & 3,308    & 86,067                                     \\
2nd Stage                  & 14,921   & 6,095     & 7,806    & 1,761 & 10,587       & 1,773    & 43,033                                     \\
3rd Stage                  & 4,996     & 2,052      & 2,707     & 544   & 3,485         & 562      & 14,346                                     \\
Total                    & 49,999   & 20,518    & 26,155   & 5,821 & 35,370       & 5,643    & 143,446    \\      
\bottomrule
\end{tabular}}
\vskip -0.1in
\end{table}

Building on this decomposition, we leverage advanced proprietary MLLMs to synthesize high-quality structured CoTs conditioned on both the image and its label. Concretely, we design a dedicated prompt template that specifies the required CoT format and criteria, and apply it to image–label pairs from the VECBench training split. The proprietary MLLM (Seed1.5-VL \citep{seed15-vl}) then generates CoTs that adhere to the prescribed structure. To mitigate potential errors introduced during synthesis, we incorporate an additional filtering stage, which leverages another MLLM (GPT-5 \citep{gpt5}) to evaluate each synthesized CoT for issues such as hallucinations, label leakage, and logical inconsistencies. Only CoTs that pass this rigorous screening are retained as high‑quality outputs, ultimately forming triplets together with their original image–label pairs. This procedure leads to the VEC-CoT dataset, which contains 143,446 image–label–CoT triplets spanning diverse data sources and two major VEC task families. Detailed statistics are reported in \cref{m:tab1}. We subsequently split the dataset into three disjoint subsets, following a 6:3:1 ratio, denoted VEC-CoT-1/2/3, corresponding to the utilization across the three training stages.

Let VEC-CoT-1 be denoted as $\{(x_i, y_i^*)\}^N_{i=1}$, consisting of $N$ queries paired with high-quality responses. Each query $x_i$ contains an image and a question, while the response comprises a CoT enclosed within a \texttt{<think>} tag pair and an emotion label enclosed within an \texttt{<answer>} tag pair. Each step in the CoT is further structured by tag pairs corresponding to its semantic role, namely, \texttt{<element>}, \texttt{<human>}, \texttt{<context>}, \texttt{<interaction>}, and \texttt{<analysis>}. In general, the response $y_i^*$ is an $n$-length text sequence, which we denote as $y_i^*=[y_{i,1}^*, y_{i,2}^*,\cdots, y_{i,n}^*]$. We adopt Qwen2.5-VL (7B) \citep{qwen25-vl} as the base model, denoted by $\pi_{\theta}$, and perform SFT with a cross-entropy objective. It encourages the model to learn reasoning paths from the synthesized CoTs, endowing it with structured reasoning that mirrors human cognition:

\vskip -0.1in
\begin{equation}
\mathcal{L}_{\text{stage1}}(\theta)=-\mathbb{E}_t\Big[\log \pi_\theta(y_{i,t}^*|x_i,y_{i,<t}^*)\Big]\text{, }\mathbb{E}_t: \text{expectation across variable } t.
\end{equation}

\subsection{Stage 2: Teaching to Verbalize Confidence}
\label{subsec:3_2}

\begin{figure}[t]
    \centering
    \includegraphics[width=1\linewidth]{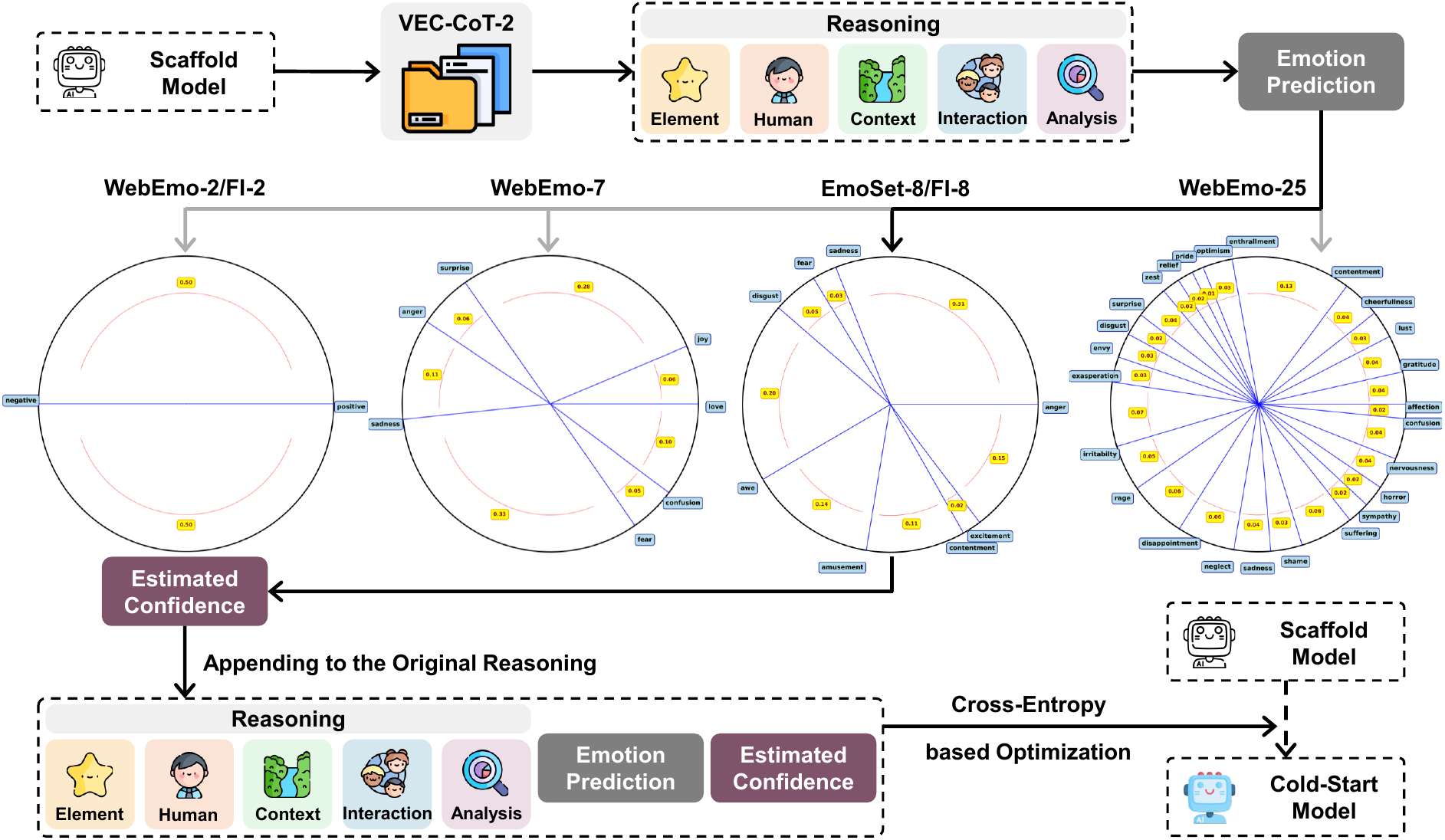}
    \caption{Illustration of the 2nd training stage. The scaffold model first performs inference on unseen data, producing a CoT and an emotion prediction. This prediction is then mapped onto a VAD lexicon-based emotion loop to measure its normalized distance from the ground-truth label. The estimated confidence score, derived from this distance, is directly appended to the original CoT and prediction, forming the supervision data used for SFT.}
    \vskip -0.1in
    \label{fig:m5}
\end{figure}

We refer to the resulting model as the scaffold model. Although it is equipped with structured reasoning, framing VEC as a deterministic task still raises practical concerns regarding reliability. Therefore, in this stage, we further teach the model to verbalize its confidence level, as illustrated in \cref{fig:m5}. For a query $x_i$ from VEC-CoT-2, we first prompt the model to generate a response $y_i$ and extract the content within the \texttt{<answer>} tag pair as its emotion prediction. Since the model has not encountered these samples during the first stage, some predictions may disagree with the annotated labels. Given the strong semantic correlation among emotion categories, we propose quantifying human prior knowledge about such deviations as a reference for the model’s confidence estimation. For example, for an image labeled as amusement, the model should assign higher confidence estimation to an erroneous prediction of contentment than to one of anger.

To quantify this, we leverage the NRC VAD lexicon \citep{nrc-vad}, which characterizes words along three well-established affective dimensions, Valence, Arousal, and Dominance (VAD). The Euclidean distance in the VAD space thereby becomes an intuitive and psychology-grounded measurement. To normalize these distances, we design a strategy that maps the VAD representations onto a 2D circle while preserving their relative spatial relationships. For a set of flat-structured categories, we compute the pairwise VAD distances and determine the optimal loop connecting them. This ring-shaped structure allows us to obtain a perimeter-based normalized semantic distance between any two emotion categories by measuring their separation along its circumference. In cases where emotion categorization is hierarchical (i.e., Parrott's model \citep{parrott2001emotions} of WebEmo), we put a constraint on loop construction to ensure the adjacency of subcategories sharing the same parent, while keeping the rest of the procedure unchanged. The four emotion loops obtained through this procedure are shown in \cref{fig:m5}, fulfilling all subtask requirements in VEC-CoT-2. We denote the normalized semantic discrepancy between predicted emotion $e_i$ and annotated emotion $e^*_i$ as $d(e_i,e^*_i) \in [0,0.5]$.

Beyond this semantic prior, we further incorporate a probability-based prior. Given an $n$-length response sequence $y_i=[y_{i,1}, y_{i,2},\cdots, y_{i,n}]$, we take the predicted token probabilities $[p_{i,1}, p_{i,2},\cdots, p_{i,n}]$ derived from the model logits and average them over the sequence. Subsequently, we adopt a heuristic computation to unify them into a final confidence estimation $c_i^*$ of response $y_i$:

\vskip -0.1in
\begin{equation}
c_i^* = \left\{
\begin{array}{ll}
0.5 * (1 + \mathbb{E}_j[p_{i,j}]) & \text{if } e_i = e^*_i, \\
0.5 - d(e_i, e^*_i) * \mathbb{E}_j[p_{i,j}] & \text{if } e_i \neq e^*_i.
\end{array}
\right.
\end{equation}

This computation guarantees that $c_i^* \in [0.5,1.0]$ for correct predictions and $c_i^* \in [0,0.5]$ for incorrect ones. We then round this value to two decimal places and embed it within a \texttt{<confidence>} tag pair, which is appended directly after response $y_i$. Finally, the model $\pi_\theta$ is still optimized by the cross-entropy objective, where the encapsulated confidence is simplified as $c_i^*$:

\vskip -0.1in
\begin{equation}
\mathcal{L}_{\text{stage2}}(\theta)=-\mathbb{E}_t\Big[\log\pi_\theta(c_{i,t}^*|x_i,y_i,c_{i,<t}^*)\Big].
\end{equation}

\subsection{Stage 3: Calibrating Confidence Expression}
\label{subsec:3_3}

We refer to the resulting model as the cold-start model. In this stage, we further optimize both confidence estimation and emotion prediction through RL. For a query $x_i$ from VEC-CoT-3, we denote the model's response as $y_i$. The emotion prediction $e_i$ is extracted from the \texttt{<answer>} tag pair, and the confidence estimate $c_i$ is extracted from the \texttt{<confidence>} tag pair. Based on empirical exploration, we adopt the log-likelihood reward for confidence calibration. Let $\mathbb{I}(e_i)$ denote a binary indicator (0/1) of whether the prediction is correct. This reward is formulated as:

\vskip -0.1in
\begin{equation}
\label{eq4}
\mathcal{R}_{\text{conf}}(y_i)=\mathbb{I}(e_i)*log(c_i)+(1-\mathbb{I}(e_i))*log(1-c_i).
\end{equation}

It encourages high confidence for correct predictions and low confidence for incorrect ones, while severely penalizing overconfident cases where confidence is misaligned with prediction correctness. For correctness, we adopt a vanilla outcome-based reward $\mathcal{R}_{\text{correct}}(y_i) = \mathbb{I}(e_i)$. We further incorporate a binary (0/1) format reward $\mathcal{R}_{\text{format}}(y_i)$ to verify whether the model outputs properly structured reasoning and whether each component is correctly enclosed within its designated tag pair. The overall reward is their sum: $\mathcal{R}(y_i)=\mathcal{R}_{\text{conf}}(y_i)+\mathcal{R}_{\text{correct}}(y_i)+\mathcal{R}_{\text{format}}(y_i).$

Subsequently, we adopt a GRPO \citep{grpo} variant for policy optimization. Given query $x_i$, the algorithm first samples a group of $G$ responses $[y_{i,1},y_{i,2},\cdots,y_{i,G}]$ using a previous policy model $\pi_{\text{old}}$. According to \cite{grpo_overconfidence}, we compute the advantage $A_{i,g}$ of response $y_{i,g}$ without the normalization: $A_{i,g}=\mathcal{R}(y_{i,g})-\text{mean}(\mathcal{R})$. The optimization objective of the cold-start model $\pi_\theta$ in this stage is formulated as:

\vskip -0.1in
\begin{equation}
\mathcal{L}_{\text{stage3}}(\theta)=-\mathbb{E}_g\Big[ \mathbb{E}_t\Big[\min(\rho_{i,g,t}(\theta)A_{i,g},\text{clip}(\rho_{i,g,t}(\theta),1\pm\epsilon)A_{i,g})-\beta D_\text{KL}(\pi_\theta||\pi_{\text{ref}}) \Big]\Big].
\end{equation}

$\rho_{i,g,t}(\theta)=\pi_\theta(y_{i,g,t}|x_i,y_{i,g,<t})/\pi_{\text{old}}(y_{i,g,t}|x_i,y_{i,g,<t})$ serves as a correction factor to simulate on-policy sampled distribution. The KL-divergence term aims to constrain the model within a trusted region around the reference model $\pi_{\text{ref}}$.

\section{Experiments}
\label{sec:exp}

To validate the effectiveness of the proposed method, we conduct comprehensive experiments in this section. Experimental preparations are described in \cref{subsec:4_1}, \cref{subsec:4_2}, and \cref{subsec:4_3}. The main performance comparisons are reported in \cref{subsec:4_4} and \cref{subsec:4_5}, while analytical studies are presented in \cref{subsec:4_6}, \cref{subsec:4_7}, \cref{subsec:4_8}, and \cref{subsec:4_9}.

\subsection{Implementation Details}
\label{subsec:4_1}
In the first stage, we adopt a batch size of 8, freeze the vision encoder, and fully fine-tune the LLM with a learning rate of 1e-5. In the second stage, we lower the learning rate to 4e-6 while keeping all other configurations unchanged. In the third stage, we set the KL coefficient to 0.01 and fine-tune all parameters (including the vision encoder) under a learning rate of 1e-6.

\subsection{Evaluation on VECBench}
\label{subsec:4_2}
The strong generalizability of MLLMs allows them to operate beyond fixed train–test dataset pairs. While this flexibility is desirable, it inevitably raises concerns about fairness during evaluation. This issue deserves particular attention in VEC, which is composed of diversified yet fragmented datasets, where data used to train one model may inadvertently appear in the test split of another. To establish a unified and fair evaluation protocol, we consolidate a set of popular datasets into VECBench. Based on it, we designate a held-out portion of data exclusively for testing, mitigating potential data leakage.

Specifically, VECBench unifies six datasets. \textbf{1). FI} \citep{fi2016} is collected from Flickr and Instagram and annotated via crowdsourcing using Mikels’ eight emotions \citep{mikels8}. In line with previous implementations, we map half of its labels to binary emotion polarity (FI-2), and retain the remaining half as FI-8. From each subset, we sample 1,000 instances for testing.  \textbf{2). WebEmo} \citep{webemo2018} is a large-scale dataset built upon keyword retrieval, annotated according to Parrott’s hierarchical model. Each image receives a fine-grained 25-class label, which can be further grouped into 7 and subsequently 2 higher-level classes. This hierarchical structure provides a unique advantage in evaluating MLLMs in distinguishing subtle emotional differences. Following a 1:1:1 ratio across the three hierarchies, we construct WebEmo-25, WebEmo-7, and WebEmo-2, sampling 1,000 images from each subset for testing. \textbf{3). EmoSet} \citep{emoset2023} is a recent large-scale dataset also annotated with Mikels’ eight emotions, characterized by its high quality and coverage. We use it directly as EmoSet-8 and sample 1,000 instances for testing. The training split of VECBench is constructed by integrating all previous unselected samples.

While in‑domain (ID) testing captures a model’s most direct emotion comprehension, its ability to generalize to out‑of‑domain (OOD) samples is equally indispensable in real‑world applications. With this in mind, we introduce three additional datasets from distinct domains. \textbf{4) Abstract} \citep{abstract2010} consists of 229 images composed purely of color‑texture combinations; \textbf{5) Artphoto} \citep{abstract2010} includes 806 artistic works sourced from art‑sharing platforms, both of which are annotated based on Mikel's model and absent from the training split. We directly incorporate all their samples as Abstract‑8 and Artphoto‑8 for testing. \textbf{6) UnbiasedEmo} \citep{webemo2018} is a carefully curated 6‑class dataset designed to challenge models in recognizing different emotions with the same object/scene. We sample 1,000 instances from it (UnbiasedEmo‑6) for evaluation. Consequently, we group all datasets into the two mainstream VEC tasks, VER and VSA, based on their classification granularity. The resulting training split contains four VER sub-tasks (FI-8, WebEmo-7, WebEmo-25, EmoSet-8) and two VSA sub-tasks (FI-2, WebEmo-2). The test split includes the corresponding four ID VER and two ID VSA sub-tasks, as well as three OOD VER sub-tasks (Abstract-8, ArtPhoto-8, UnbiasedEmo-8). Summary statistics for all components are shown in \cref{fig:m3}.

\begin{table}[t]
\vskip -0.02in
\caption{Comparison of advanced MLLMs on emotion prediction on ID VER tasks of VECBench. \textbf{Bold} and \underline{underline} indicate the best and second-best results, respectively.}
\vskip -0.1in
\label{m:tab2}
\centering
\resizebox{1\linewidth}{!}{
\renewcommand{\arraystretch}{1.1}
\begin{tabular}{lcccccccccc}
\toprule
\multirow{2}{*}{\bf ID VER} & \multicolumn{2}{c}{FI-8} & \multicolumn{2}{c}{WebEmo-7} & \multicolumn{2}{c}{WebEmo-25} & \multicolumn{2}{c}{EmoSet-8} & \multicolumn{2}{c}{Average} \\
\cmidrule(l){2-3} \cmidrule(l){4-5} \cmidrule(l){6-7} \cmidrule(l){8-9} \cmidrule(l){10-11}
                        & Acc $\uparrow$        & F1 $\uparrow$         & Acc $\uparrow$          & F1 $\uparrow$          & Acc $\uparrow$          & F1 $\uparrow$           & Acc $\uparrow$          & F1 $\uparrow$          & Acc $\uparrow$         & F1 $\uparrow$          \\
\midrule
\multicolumn{11}{c}{\cellcolor[HTML]{ebebeb}Open-Source MLLM} \\
Qwen2.5-VL \citep{qwen25-vl}        & 51.90       & 50.95      & 43.60         & 43.87        & 21.00         & 21.11         & 47.30         & 45.01        & 40.73        & 40.24        \\
InternVL3.5 \citep{internvl35}            & 53.10       & 51.96      & 42.00         & 41.09        & 16.10         & 15.46         & 51.90         & 50.47        & 40.78        & 39.75        \\
Qwen3-VL  \citep{qwen3vl}              & 56.50       & 54.57      & \underline{49.70}         & \underline{49.56}        & \underline{21.10}         & \underline{21.72}         & 52.60         & 50.64        & \underline{44.98}        & 44.12        \\
\multicolumn{11}{c}{\cellcolor[HTML]{ebebeb}Proprietary MLLM} \\
GPT-5 \citep{gpt5}                  & 53.80       & 53.06      & 44.50         & 45.61        & 19.50         & 19.37         & 54.80         & 52.94        & 43.15        & 42.75        \\
\multicolumn{11}{c}{\cellcolor[HTML]{ebebeb}Emotion-Oriented MLLM} \\
EmoVIT \citep{emovit2024}                 & \underline{59.70}       & \underline{58.16}      & 44.40         & 48.34        & 15.00         & 15.12         & \underline{59.40}         & \underline{59.17}        & 44.63        & \underline{45.20}        \\
Emotion-LLaMA \citep{emotion-llama}     & 23.50       & 18.97      & 18.70         & 20.11        & 10.30         & 10.04         & 18.70         & 13.26        & 17.80        & 15.60        \\
\rowcolor[HTML]{e1e2ff}EmoCaliber (\textbf{Ours})              & \textbf{69.70}     & \textbf{70.00}      & \textbf{54.00}         & \textbf{52.81}        & \textbf{28.30}         & \textbf{24.28}         & \textbf{68.10}         & \textbf{67.72}        & \textbf{55.03}        & \textbf{53.70}       \\
\bottomrule
\end{tabular}}
\vskip -0.1in
\end{table}

\begin{table}[t]
\vskip -0.02in
\caption{Comparison of MLLMs on emotion prediction on ID VSA tasks of VECBench.}
\vskip -0.1in
\label{m:tab3}
\centering
\resizebox{0.73\linewidth}{!}{
\renewcommand{\arraystretch}{1.1}
\begin{tabular}{lcccccc}
\toprule
\multirow{2}{*}{\bf ID VSA} & \multicolumn{2}{c}{FI-2} & \multicolumn{2}{c}{WebEmo-2} & \multicolumn{2}{c}{Average} \\
\cmidrule(l){2-3} \cmidrule(l){4-5} \cmidrule(l){6-7}
                            & Acc $\uparrow$       & F1 $\uparrow$         & Acc $\uparrow$         & F1 $\uparrow$           & Acc $\uparrow$         & F1 $\uparrow$         \\
\midrule
\multicolumn{7}{c}{\cellcolor[HTML]{ebebeb}Open-Source MLLM} \\
Qwen2.5-VL \citep{qwen25-vl}                 & 84.90       & 84.82       & 69.50         & 69.64         & 77.20        & 77.23       \\
InternVL3.5  \citep{internvl35}               & 85.60       & 85.30        & 70.30         & 69.47         & 77.95        & 77.39       \\
Qwen3-VL \citep{qwen3vl}                   & 87.10       & 87.04       & \textbf{75.90}         & \underline{75.71}         & \underline{81.50}        & \underline{81.38}       \\
\multicolumn{7}{c}{\cellcolor[HTML]{ebebeb}Proprietary MLLM} \\
GPT-5  \citep{gpt5}                     & \textbf{88.60}       & \textbf{88.40}        & 73.70         & 73.51         & 81.15        & 80.96       \\
\multicolumn{7}{c}{\cellcolor[HTML]{ebebeb}Emotion-Oriented MLLM} \\
EmoVIT \citep{emovit2024}                     & 85.50       & 85.60        & 74.40         & 74.67         & 79.95        & 80.14       \\
Emotion-LLaMA \citep{emotion-llama}              & 56.40       & 63.15       & 44.90         & 48.21         & 50.65        & 55.68       \\
\rowcolor[HTML]{e1e2ff}EmoCaliber (\textbf{Ours})                  & \underline{88.10}       & \underline{88.16}       & \underline{75.80}         & \textbf{75.78}         & \textbf{81.95}        & \textbf{81.97}    \\
\bottomrule
\end{tabular}}
\vskip -0.1in
\end{table}

\subsection{Baselines and Evaluation Metrics}
\label{subsec:4_3}
To ensure fair and meaningful comparisons, our baselines include state-of-the-art open-source MLLMs (\textbf{Qwen2.5-VL} \citep{qwen25-vl}, \textbf{InternVL3.5} \citep{internvl35}, and \textbf{Qwen3-VL} \citep{qwen3vl}), as well as a representative proprietary MLLM (\textbf{GPT-5} \citep{gpt5}). For MLLMs that have undergone emotion-oriented optimization, we include \textbf{EmoVIT} \citep{emovit2024} and \textbf{Emotion-LLaMA} \citep{emotion-llama}. Notably, the reported EmoVIT results are a reproduced version based on Qwen2.5-VL. Most other emotion-oriented MLLMs are either not publicly available or not aligned with our task setting. One exception is AffectGPT \citep{affectgpt}, which is similar to Emotion-LLaMA and tailored for video input. Preliminary experiments indicate that AffectGPT performs comparably to Emotion-LLaMA, yet substantially underperforms the image-targeted baselines on VEC. For clarity and conciseness, we therefore exclude it from the main comparisons.

For emotion prediction, we report Accuracy (\textbf{Acc}) and macro-F1 \textbf{(F1)}. Accuracy measures the overall proportion of correctly classified samples, while macro-F1 emphasizes balanced performance across all categories by averaging F1 scores equally. For confidence evaluation, we follow prior work and adopt three complementary metrics. Expected Calibration Error (\textbf{ECE}) groups predictions into confidence bins and computes the discrepancy between average confidence and empirical correctness; lower values indicate better calibration. We use 10 bins, each covering a confidence interval of 0.1. The \textbf{Brier} score measures the mean squared error between predicted confidence and binary correctness, with lower values preferable. Area under ROC curve (\textbf{AUC}) evaluates the model’s ability to distinguish correct from incorrect predictions across varying confidence thresholds, with 0.5 indicating random performance and 1.0 representing ideal discrimination.

\begin{table}[t]
\vskip -0.02in
\caption{Comparison of MLLMs on emotion prediction on OOD VER tasks of VECBench.}
\vskip -0.1in
\label{m:tab4}
\centering
\resizebox{0.9\linewidth}{!}{
\renewcommand{\arraystretch}{1.1}
\begin{tabular}{lcccccccc}
\toprule
\multirow{2}{*}{\bf OOD VER} & \multicolumn{2}{c}{UnbiasedEmo-6} & \multicolumn{2}{c}{Abstract-8} & \multicolumn{2}{c}{Artphoto-8} & \multicolumn{2}{c}{Average} \\
\cmidrule(l){2-3} \cmidrule(l){4-5} \cmidrule(l){6-7} \cmidrule(l){8-9}
                             & Acc $\uparrow$            & F1 $\uparrow$             & Acc $\uparrow$           & F1 $\uparrow$           & Acc $\uparrow$           & F1 $\uparrow$           & Acc $\uparrow$         & F1 $\uparrow$          \\
\midrule
\multicolumn{9}{c}{\cellcolor[HTML]{ebebeb}Open-Source MLLM} \\
Qwen2.5-VL \citep{qwen25-vl}                  & 77.80           & 77.76           & 28.38          & \underline{31.35}         & 36.97          & 37.43         & 47.72        & 48.85        \\
InternVL3.5 \citep{internvl35}                 & 76.80           & 76.66           & 27.95          & 28.63         & 40.82          & 41.80         & 48.52        & 49.03        \\
Qwen3-VL \citep{qwen3vl}                    & \textbf{85.40}           & \textbf{85.47}           & 23.58          & 22.87         & 40.32          & 41.10         & 49.77        & 49.81        \\
\multicolumn{9}{c}{\cellcolor[HTML]{ebebeb}Proprietary MLLM} \\
GPT-5 \citep{gpt5}                       & 64.20           & 63.02           & \textbf{41.92}          & \textbf{43.48}         & \textbf{45.53}          & \textbf{45.33}         & \textbf{50.55}        & 50.61        \\
\multicolumn{9}{c}{\cellcolor[HTML]{ebebeb}Emotion-Oriented MLLM} \\
EmoVIT \citep{emovit2024}                       & 78.50           & 78.24           & 28.81          & 28.80         & 41.61          & 38.09         & 49.64        & 48.38        \\
Emotion-LLaMA \citep{emotion-llama}               & 40.50           & 46.41           & 10.48          & 7.78          & 22.95          & 16.46         & 24.64        & 23.55        \\
\rowcolor[HTML]{e1e2ff}EmoCaliber (\textbf{Ours})                   & \underline{79.90}           & \underline{80.41}           & \underline{29.26}          & 29.94         & \underline{41.94}          & \underline{41.84}         & \underline{50.37}        & \textbf{50.73}   \\
\bottomrule
\end{tabular}}
\vskip -0.1in
\end{table}

\subsection{Comparison on Emotion Prediction}
\label{subsec:4_4}
\cref{m:tab2}, \cref{m:tab3}, and \cref{m:tab4} report the emotion prediction performance of MLLMs on the ID VER, ID VSA, and OOD VER tasks of VECBench, respectively. On the ID VER tasks, our proposed EmoCaliber significantly outperforms all competitors, including advanced open-source, proprietary, and emotion-oriented MLLMs. In particular, EmoCaliber achieves an average accuracy of 53.70, exceeding the second-best model, Qwen3-VL, by 10.05. These results validate the effectiveness of our structured reasoning design. By mirroring human affective cognition, it enables the model to capture fine-grained visual cues and more accurately model affective correlations, thereby leading to proficient emotion comprehension. It is worth noting that Emotion-LLaMA exhibits substantially lower performance compared to the others. As it is originally designed for video input, we include it as a representative baseline to highlight the necessity of image-oriented methods for VEC.

On the ID VSA tasks, although the performance gains are less pronounced, EmoCaliber still achieves the highest average accuracy and ranks within the top two across all subtasks. We attribute this to the relatively simpler nature of VSA, where the advantages of stronger affective reasoning may be partially offset by the general comprehension capabilities of more advanced models. 

On the OOD VER tasks, EmoCaliber also delivers competitive performance with an average accuracy of 50.37, the second‑highest among all models and only slightly behind GPT‑5 (50.55). Since these test samples lie entirely outside the training domains, this result validates that the gains of our model are not merely due to memorizing ID patterns. Instead, EmoCaliber generalizes robustly to more diverse data, achieving emotion prediction performance comparable to proprietary models.

\begin{table}[t]
\vskip -0.02in
\caption{Comparison of MLLMs on confidence estimation on VECBench. ``Verb'' denotes using the verbalized confidence; ``Ans'' denotes using the token probability at the answer position; ``Avg'' denotes using the average probability over all response tokens.}
\vskip -0.1in
\label{m:tab5}
\centering
\resizebox{1\linewidth}{!}{
\renewcommand{\arraystretch}{1.1}
\begin{tabular}{lcccccccccc}
\toprule
\multicolumn{2}{l}{\multirow{2}{*}{\bf Confidence Estimation}}  &  \multicolumn{3}{c}{ID VER} & \multicolumn{3}{c}{ID VSA} & \multicolumn{3}{c}{OOD VER} \\
\cmidrule(l){3-5} \cmidrule(l){6-8} \cmidrule(l){9-11}
\multicolumn{2}{l}{\multirow{2}{*}{}}              & ECE $\downarrow$    & Brier $\downarrow$  & AUC $\uparrow$   & ECE $\downarrow$    & Brier $\downarrow$  & AUC $\uparrow$   & ECE $\downarrow$    & Brier $\downarrow$  & AUC $\uparrow$    \\
\midrule
\multicolumn{11}{c}{\cellcolor[HTML]{ebebeb}Open-Source MLLM} \\
\multirow{3}{*}{Qwen2.5-VL \citep{qwen25-vl}}      & Verb                     & 49.20   & 48.18   & 56.21  & 13.81   & 17.03   & 59.35  & 34.15   & 34.93   & 60.93   \\
                                 & Ans                      & 50.61   & 51.06   & 59.93  & 16.85   & 18.56   & 61.69  & 35.23   & 38.58   & 60.55   \\
                                 & Avg                      & 42.44   & 41.98   & 58.58  & 2.94    & 16.06   & 64.30  & 25.59   & 29.90   & 67.57   \\
\arrayrulecolor{gray} \cline{1-11}
\multirow{2}{*}{InternVL3.5 \citep{internvl35}}     & Ans                      & 48.89   & 52.82   & 56.30  & 27.12   & 29.55   & 67.09  & 36.69   & 39.93   & 47.50   \\
                                 & Avg                      & 39.80   & 39.50   & 59.24  & \underline{2.47}    & 16.88   & 60.49  & 23.33   & 29.53   & 57.81   \\
\arrayrulecolor{gray} \cline{1-11}
\multirow{3}{*}{Qwen3-VL \citep{qwen3vl}}        & Verb                     & 48.95   & 47.21   & 65.60  & 12.13   & 15.24   & \underline{67.54}  & 30.05   & 30.68   & 69.47   \\
                                 & Ans                      & \underline{26.31}   & 39.62   & 53.13  & \textbf{0.37}    & 17.77   & 51.42  & 21.95   & 32.38   & 42.35   \\
                                 & Avg                      & 46.05   & 45.71   & 59.87  & 11.18   & 16.14   & 60.75  & 30.66   & 32.73   & 69.99   \\
\arrayrulecolor{gray} \cline{1-11}
\multicolumn{11}{c}{\cellcolor[HTML]{ebebeb}Proprietary MLLM} \\
GPT5 \citep{gpt5}                            & Verb                     & 28.20  & \underline{29.99}   & \underline{69.43}  & 3.39    & \textbf{13.69}   & \textbf{74.28}  & \underline{19.35}   & \underline{25.82}   & \underline{70.92}   \\
\arrayrulecolor{gray} \cline{1-11}
\multicolumn{11}{c}{\cellcolor[HTML]{ebebeb}Emotion-Oriented MLLM} \\
\multirow{2}{*}{EmoVIT \citep{emovit2024}}          & Ans                      & 33.09   & 39.11   & 49.63  & 16.27   & 17.85   & 49.76  & 22.70   & 29.91   & 57.43   \\
                                 & Avg                      & 33.52   & 35.91   & 57.43  & 3.62    & 15.01   & 62.81  & 24.47   & 29.45   & 67.29   \\
\arrayrulecolor{gray} \cline{1-11}
Emotion-LLaMA \citep{emotion-llama}                   & Avg                      & 62.19   & 53.31   & 50.00  & 29.35   & 33.61   & 50.00  & 49.83   & 45.90   & 50.00   \\
\arrayrulecolor{gray} \cline{1-11}
\rowcolor[HTML]{e1e2ff}EmoCaliber (\textbf{Ours})                 & Verb & \textbf{13.63}   & \textbf{22.77}   & \textbf{70.90}  & 4.76    & \underline{14.68}   & 66.09  & \textbf{12.17}   & \textbf{22.41}   & \textbf{72.17}  \\
\arrayrulecolor{black} \bottomrule
\end{tabular}}
\vskip -0.1in
\end{table}

\begin{table}[t]
\vskip -0.02in
\caption{Evolvement of model capability along the three training stages. For fair comparison, we uniformly adopt verbalized confidence as the model's estimated confidence.}
\vskip -0.1in
\label{m:tab6}
\centering
\resizebox{0.86\linewidth}{!}{
\renewcommand{\arraystretch}{1.1}
\begin{tabular}{lccccccccc}
\toprule
\multirow{2}{*}{\bf Evolvement} & \multicolumn{3}{c}{ID VER} & \multicolumn{3}{c}{ID VSA} & \multicolumn{3}{c}{OOD VER} \\
\cmidrule(l){2-4} \cmidrule(l){5-7} \cmidrule(l){8-10}
                  & Acc $\uparrow$    & ECE $\downarrow$    & Brier $\downarrow$ & Acc $\uparrow$    & ECE $\downarrow$   & Brier $\downarrow$  & Acc $\uparrow$    & ECE  $\downarrow$   & Brier $\downarrow$  \\
\midrule
\multicolumn{10}{c}{\cellcolor[HTML]{ebebeb}Initialized from Qwen2.5-VL} \\
Base Model        & 40.73   & -       & -      & 77.20   & -      & -       & 47.72   & -       & -       \\
\multicolumn{10}{c}{\cellcolor[HTML]{ebebeb}Stage 1: Endowing with Structured Reasoning} \\
Scarffold Model   & 54.47   & -       & -      & 81.75   & -      & -       & 50.04   & -       & -       \\
\multicolumn{10}{c}{\cellcolor[HTML]{ebebeb}Stage 2: Teaching to Verbalize Confidence} \\
Cold-Start Model  & 54.04   & 14.67   & 23.87  & 81.60   & 7.28   & 14.79   & 49.81   & 13.01   & 22.94   \\
\multicolumn{10}{c}{\cellcolor[HTML]{ebebeb}Stage 3: Calibrating Confidence Expression} \\
EmoCaliber        & \textbf{55.03}   & \textbf{13.63}   & \textbf{22.77}  & \textbf{81.95}   & \textbf{4.76}   & \textbf{14.68}   & \textbf{50.37}   & \textbf{12.17}   & \textbf{22.41}  \\
\bottomrule
\end{tabular}}
\vskip -0.1in
\end{table}

\subsection{Comparison on Confidence Estimation}
\label{subsec:4_5}
\cref{m:tab5} compares the confidence estimation performance of MLLMs on VECBench. For each baseline, we elicit confidence estimates using up to three strategies when applicable. ``Verb'' denotes explicitly prompting the model to output a confidence score; ``Ans'' extracts the predicted probability of the first decisive token in the answer; and ``Avg'' computes the average token probability over the entire output sequence, including both the reasoning and the final answer.

Overall, EmoCaliber significantly outperforms all competing models on both ID VER and OOD VER tasks. Notably, on ID VER, EmoCaliber achieves an ECE of 13.63, which is nearly half that of the second-best baseline, Qwen3-VL (Ans) at 26.31. A similar performance gap is observed on OOD VER, highlighting EmoCaliber’s clear advantage in calibrated confidence estimation for VER tasks. On ID VSA, EmoCaliber performs slightly below GPT-5 while remaining broadly comparable to other state-of-the-art MLLMs. We attribute this behavior to the training strategy, as it is inherently challenging to simultaneously prevent overconfidence in fine-grained multi-class tasks and ensure sufficient confidence in coarse-grained binary tasks. We further analyze this trade-off in the reward ablation study. Collectively, EmoCaliber demonstrates strong confidence estimation capabilities, establishing a solid, practical baseline for building reliable VEC systems.

\subsection{Analysis of Model Evolvement}
\label{subsec:4_6}
\cref{m:tab6} compares emotion prediction and confidence estimation performance before and after each training stage. As shown, the first stage training substantially improves emotion prediction accuracy across all tasks. This confirms the consistent benefits of structured affective reasoning across different data domains and task granularities. In the second stage, the model is taught to express confidence, which is accompanied by a slight degradation in emotion prediction performance. Finally, in the third stage, both emotion prediction and confidence estimation performances improve simultaneously. The former not only reverses the decline observed in the second stage but also surpasses the scaffold model, while the latter shows consistent gains across all confidence metrics. These results validate the effectiveness of RL in calibrating confidence expression. Overall, the results demonstrate that each training stage fulfills its intended objective, jointly contributing to the final performance of EmoCaliber.

\begin{table}[t]
\vskip -0.02in
\caption{Influence of different learning rate settings during the first stage training.}
\vskip -0.1in
\label{m:tab7}
\centering
\resizebox{0.86\linewidth}{!}{
\renewcommand{\arraystretch}{1.1}
\begin{tabular}{lccccccccc}
\toprule
\multirow{2}{*}{\bf 1st Stage} & Learning & \multicolumn{2}{c}{ID VER} & \multicolumn{2}{c}{ID VSA} & \multicolumn{2}{c}{OOD VER} & \multicolumn{2}{c}{Average} \\
\cmidrule(l){3-4} \cmidrule(l){5-6} \cmidrule(l){7-8} \cmidrule(l){9-10} 
                           & Rate     & Acc $\uparrow$         & F1 $\uparrow$         & Acc $\uparrow$          & F1 $\uparrow$         & Acc $\uparrow$         & F1  $\uparrow$         & Acc  $\uparrow$        & F1  $\uparrow$         \\
\midrule
\rowcolor[HTML]{ebebeb}Base Model                 & -        & 40.73        & 40.24       & 77.20        & 77.23       & 47.72        & 48.85        & 55.22        & 55.44        \\
Scaffold Model \#1         & 1e-6     & 51.54        & 50.74       & 81.65        & \underline{81.79}       & \underline{50.20}        & 50.35        & 61.13        & 60.96        \\
Scaffold Model \#2         & 2e-6     & 52.80        & 51.71       & 81.30        & 81.36       & 50.18        & 49.84        & 61.43        & 60.97        \\
Scaffold Model \#3         & 4e-6     & 53.84        & 52.94       & 81.15        & 81.23       & 49.51        & 49.66        & 61.50        & 61.28        \\
Scaffold Model \#4         & 8e-6     & 52.84        & 51.72       & 81.45        & 81.50       & \textbf{50.22}        & \textbf{50.84}        & 61.50        & 61.35        \\
\rowcolor[HTML]{e1e2ff}Scaffold Model \#5         & 1e-5     & \textbf{54.47}      & \textbf{53.58}       & \underline{81.75}       & 81.78       & 50.04        & \underline{50.55}        & \textbf{62.09}        & \textbf{61.97}        \\
Scaffold Model \#6         & 2e-5     & 53.17        & 52.23       & \textbf{81.95}        & \textbf{81.99}       & 49.92        & 50.11        & \underline{61.68}        & \underline{61.44}        \\
Scaffold Model \#7         & 4e-5     & 53.17        & 52.31       & 81.30        & 81.30       & 49.20        & 50.30        & 61.22        & 61.30        \\
Scaffold Model \#8         & 8e-5     & \underline{53.87}       & \underline{53.10}       & 80.15        & 80.12       & 47.38        & 48.15        & 60.47        & 60.46    \\
\bottomrule
\end{tabular}}
\vskip -0.1in
\end{table}

\begin{table}[t]
\vskip -0.02in
\caption{Influence of different learning rate settings during the second stage training. Bolded results indicate the best performance excluding the scaffold model.}
\vskip -0.1in
\label{m:tab8}
\centering
\resizebox{0.92\linewidth}{!}{
\renewcommand{\arraystretch}{1.1}
\begin{tabular}{lccccccccc}
\toprule
\multirow{2}{*}{\bf 2nd Stage} & Learning & \multicolumn{2}{c}{ID VER} & \multicolumn{2}{c}{ID VSA} & \multicolumn{2}{c}{OOD VER} & \multicolumn{2}{c}{Average} \\
\cmidrule(l){3-4} \cmidrule(l){5-6} \cmidrule(l){7-8} \cmidrule(l){9-10}
                           & Rate     & Acc  $\uparrow$        & Brier $\downarrow$      & Acc $\uparrow$        & Brier $\downarrow$      & Acc $\uparrow$         & Brier $\downarrow$       & Acc  $\uparrow$        & Brier $\downarrow$       \\
\midrule
\rowcolor[HTML]{ebebeb}Scarffold Model            & -        & 54.47        & -           & 81.75        & -           & 50.04        & -            & 62.09        & -            \\
Cold-Start Model \#1       & 1e-6     & 52.97        & 25.22       & \underline{81.60}        & 14.94       & 49.28        & 23.60        & 61.28        & 21.25        \\
Cold-Start Model \#2       & 2e-6     & \underline{53.27}        & 24.74       & 80.90        & 14.90       & \textbf{50.05}        & \textbf{22.88}        & 61.41        & 20.84        \\
\rowcolor[HTML]{e1e2ff}Cold-Start Model \#3       & 4e-6     & \textbf{54.05}        & \textbf{23.87}       & \underline{81.60}        & 14.79       & \underline{49.81}        & \underline{22.94}        & \textbf{61.82}        & \textbf{20.53}        \\
Cold-Start Model \#4       & 6e-6     & \textbf{54.05}        & \underline{24.56}       & \textbf{81.85}        & 14.78       & 48.55        & 23.05        & \underline{61.48}        & \underline{20.80}        \\
Cold-Start Model \#5       & 8e-6     & 52.94        & 24.71       & 81.30        & \underline{14.73}       & 49.68        & 23.24        & 61.31        & 20.89        \\
Cold-Start Model \#6       & 1e-5     & 53.22        & 24.99       & 81.25        & \textbf{14.61}       & 49.61        & 23.28        & 61.36        & 20.96       \\
\bottomrule
\end{tabular}}
\vskip -0.1in
\end{table}

\subsection{Ablation of Learning Rate}
\label{subsec:4_7}
Given the complexity of jointly optimizing the three‑stage training framework, we adopt a greedy strategy for hyperparameter tuning to pursue a local optimum. The effects of varying learning rates for the first and second training stages are presented in \cref{m:tab7} and \cref{m:tab8}, respectively.

In the first stage, the model is supervised to acquire structured reasoning capabilities. From \cref{m:tab7}, we observe that a wide range of learning rates consistently lead to stable improvements across all tasks, validating that the benefits of the structured reasoning design and constructed VEC-CoT dataset are robust to optimization settings. Nevertheless, due to the sensitivity of SFT to learning rate selection, different choices still exert a non-negligible impact on performance. Specifically, lower learning rates lead to slightly inferior performance on ID VER and ID VSA tasks, suggesting potential underfitting. In contrast, higher learning rates result in degraded performance on OOD VER tasks, possibly due to overfitting. Balancing these trade-offs, we adopt a learning rate of 1e-5 in the final implementation. Empirically, this setting achieves the highest average accuracy across tasks, providing a favorable balance between ID performance and OOD generalization.

In the second stage, the model is trained to verbalize confidence, with optimization focused on confidence-related tokens. As shown in \cref{m:tab8}, even though tokens directly responsible for emotion prediction are excluded from optimization, the resulting decline in prediction accuracy is minimal. This suggests that the optimization directions for confidence estimation and emotion prediction are not inherently conflicting, supporting their joint optimization in the subsequent RL stage. Compared to the first stage, accuracy and Brier score exhibit more complex variations across different learning rates in this stage. Taking into account overall performance, we select a learning rate of 4e‑6, which achieves top‑2 performance on most tasks. Notably, this value is lower than the 1e‑5 used in the first stage, which is intuitive given that fewer tokens are actively updated during this stage.

\begin{table}[t]
\vskip -0.02in
\caption{Influence of different RL configurations during the third stage training.}
\vskip -0.1in
\label{m:tab9}
\centering
\resizebox{1\linewidth}{!}{
\renewcommand{\arraystretch}{1.1}
\begin{tabular}{lcccccccccc}
\toprule
\multirow{2}{*}{\bf 3rd Stage} & \multirow{2}{*}{$\mathcal{R}_{\text{conf}}$} & \multirow{2}{*}{Normalize}     & \multicolumn{2}{c}{ID VER} & \multicolumn{2}{c}{ID VSA} & \multicolumn{2}{c}{OOD VER} & \multicolumn{2}{c}{Average} \\
\cmidrule(l){4-5} \cmidrule(l){6-7} \cmidrule(l){8-9} \cmidrule(l){10-11}
                                   &  &  & Acc  $\uparrow$        & Brier $\downarrow$      & Acc $\uparrow$        & Brier $\downarrow$      & Acc $\uparrow$         & Brier $\downarrow$       & Acc  $\uparrow$        & Brier $\downarrow$       \\
\midrule
\rowcolor[HTML]{ebebeb}Cold-Start Model                   & -                                  & -             & 54.05        & \underline{23.87}       & 81.60        & \underline{14.79}       & 49.81        & \underline{22.94}        & \underline{61.82}        & \underline{20.53}        \\
EmoCaliber \#1                     & Brier-based                        & \ding{51}             & 53.94        & 45.10       & 80.00        & 19.61       & \textbf{51.03}        & 40.28        & 61.66        & 35.00        \\
EmoCaliber \#2                     & Brier-based                        & \ding{55}             & \underline{54.54}        & 42.80       & \textbf{82.25}        & 16.74       & 47.33        & 41.64        & 61.37        & 33.73        \\
EmoCaliber \#3                     & Log-Likelihood                     & \ding{51}             & 53.67        & 44.55       & 81.45        & 18.07       & 49.30        & 39.77        & 61.47        & 34.13        \\
\rowcolor[HTML]{e1e2ff}EmoCaliber \#4                     & Log-Likelihood                     & \ding{55}             & \textbf{55.03}        & \textbf{22.77}       & \underline{81.95}        & \textbf{14.68}       & \underline{50.37}        & \textbf{22.41}        & \textbf{62.45}        & \textbf{20.02}     \\
\bottomrule
\end{tabular}}
\vskip -0.1in
\end{table}

\begin{figure}[t]
    \centering
    \includegraphics[width=1\linewidth]{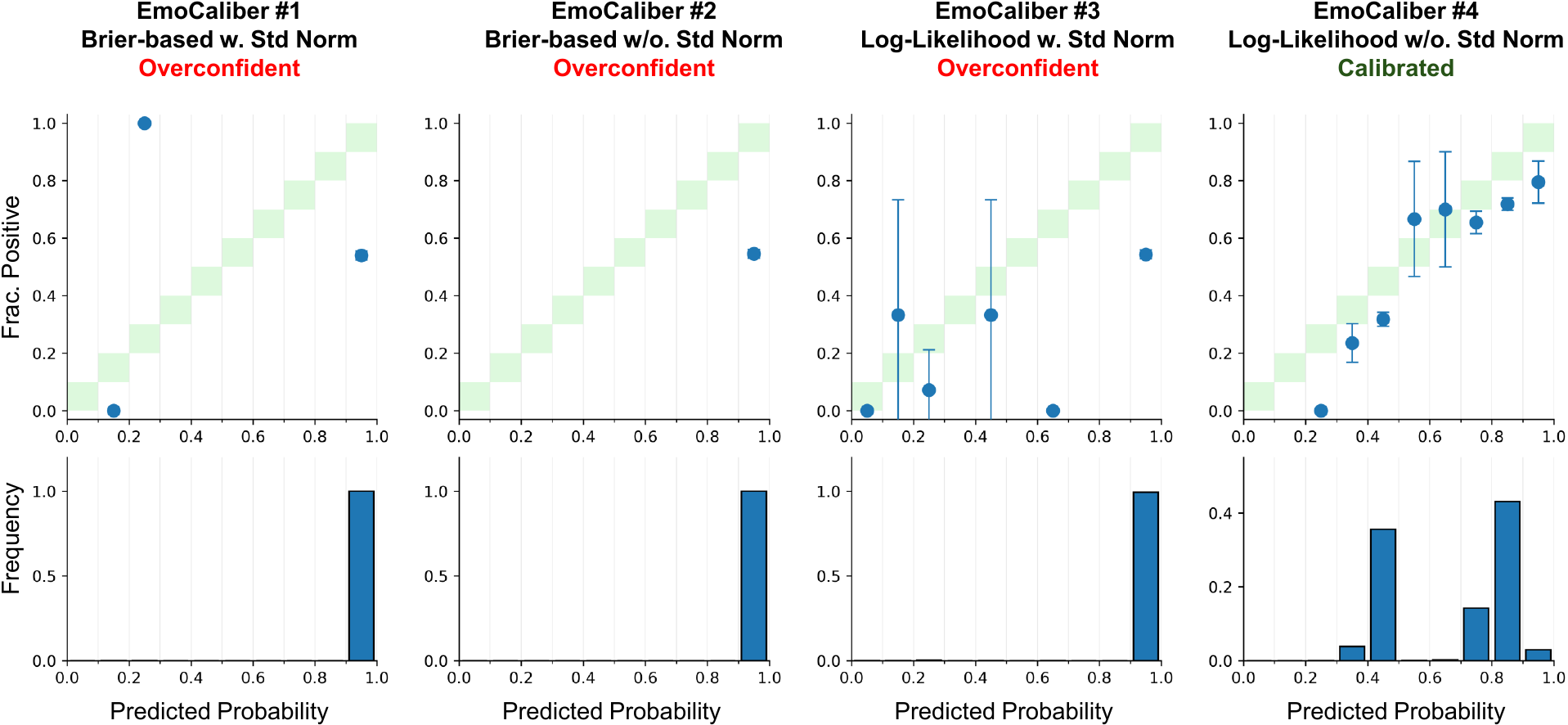}
    \caption{Distribution of verbalized confidences of models derived from different RL configurations on ID VER tasks. Error bars represent 95\% confidence intervals.}
    \vskip -0.1in
    \label{fig:m6}
\end{figure}

\subsection{Ablation of Reward Design}
\label{subsec:4_8}
Although RL has proved to be effective for confidence calibration, our experiments reveal that this process is notably unstable in VEC, with minor configuration changes leading to divergent optimization outcomes. \cref{m:tab9} compares the effects of different RL setups.

Our initial exploration (\#1) adopts the $\mathcal{R}_{\text{conf}}$ formulation from RLCR \citep{rlcr}:

\vskip -0.1in
\begin{equation}
\mathcal{R}_{\text{conf}}(y_i)=-(\mathbb{I}(e_i)-c_i) ^2,\text{ (with variable }y_i, e_i, c_i\text{ consistent with \cref{eq4})},
\end{equation}

together with the standard GRPO algorithm. While this reward provides supervision aligned with calibration metrics, it imposes relatively smooth constraints over different confidence levels. Empirically, this setting leads to severe overconfidence on VEC, with confidence estimates heavily concentrated above 0.9. A similar phenomenon has been reported in prior work \citep{grpo_overconfidence}, which attributes it to the standard advantage normalization in GRPO, potentially inducing a positive feedback loop. Specifically, when reward values within a group exhibit small variance, the normalization amplifies minor relative advantages, causing the model to repeatedly increase the probability of slightly preferred responses and drive toward extreme confidence estimations.

Motivated by this analysis, we explore removing advantage normalization and replacing the reward formulation with an alternative log-likelihood–based objective (\cref{eq4}; \#4). Acting jointly, these two modifications effectively mitigate the overconfidence issue and achieve the intended calibration behavior. In contrast, applying either modification (\#2, \#3) in isolation is insufficient. These effects are visualized in \cref{fig:m6}, where EmoCaliber exhibits well-calibrated confidence estimates.

\subsection{Case Study}
\label{subsec:4_9}
\begin{figure}[t]
    \centering
    \includegraphics[width=1\linewidth]{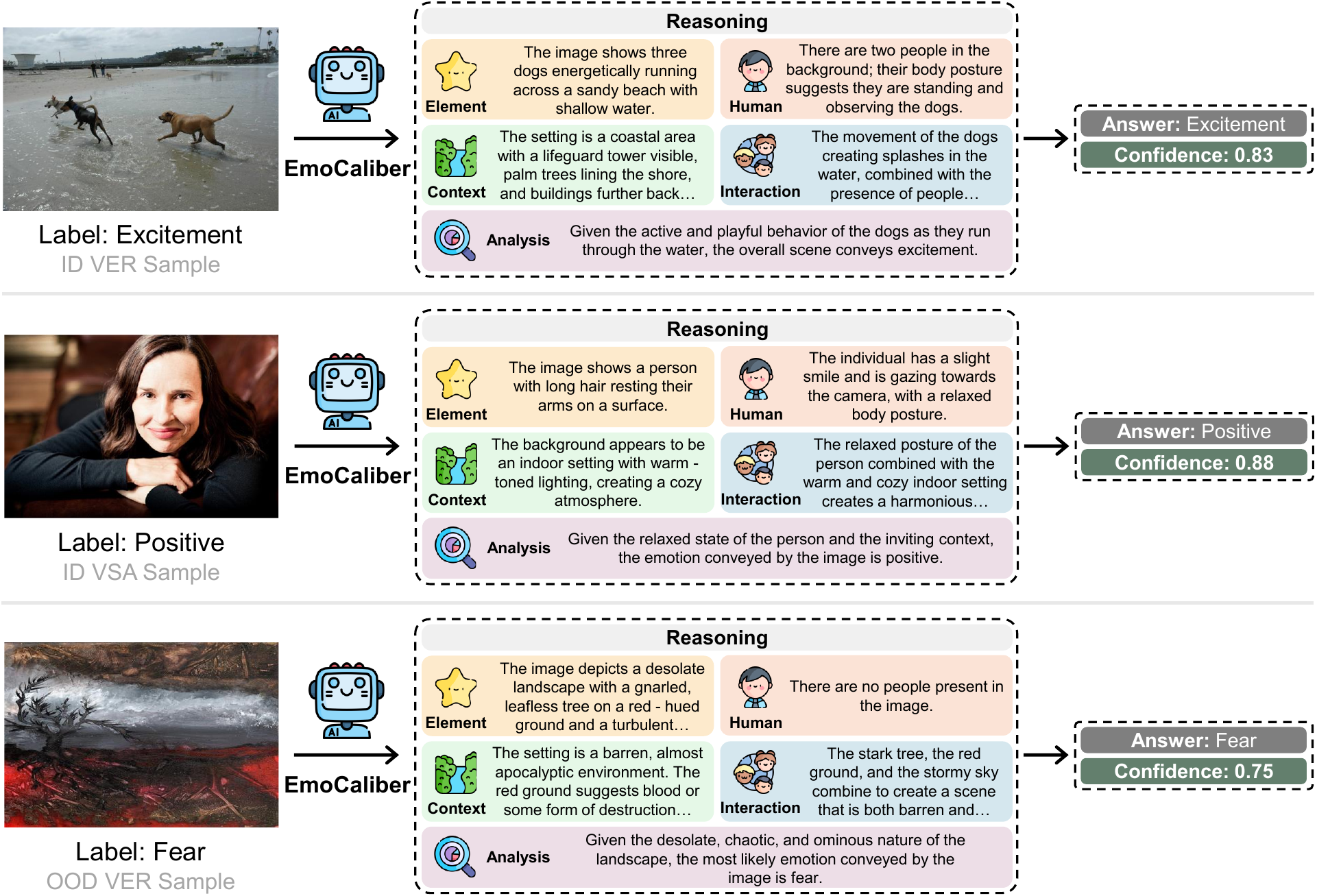}
    \caption{Representative cases of EmoCaliber performing ID VER, ID VSA, and OOD VER tasks on VECBench.}
    \vskip -0.1in
    \label{fig:m7}
\end{figure}

\cref{fig:m7} presents representative examples of EmoCaliber performing ID VER, ID VSA, and OOD VER tasks on VECBench. These cases qualitatively illustrate how EmoCaliber derives emotion predictions through structured affective reasoning, as well as how it produces explicit confidence estimates. Specifically, the structured reasoning process enables EmoCaliber to jointly consider diverse visual cues, including salient subjects (present in all cases), fine-grained objects (e.g., the two human figures in the top example), fague contextual backgrounds (such as the warm indoor atmosphere in the middle example), and artistic or stylistic elements (e.g., the apocalyptic aesthetic in the bottom example).

The confidence estimates further reflect both the inherent subjectivity of emotion perception and the model’s self-assessed reliability. EmoCaliber assigns higher confidence to relatively clear and straightforward cases (the top two examples), while expressing lower confidence for more ambiguous and uncertain inputs, such as the abstract artwork in the bottom example. Overall, these case studies provide intuitive evidence supporting the effectiveness of EmoCaliber for accurate and reliable VEC.

\section{Conclusion}
\label{sec:conclusion}
Overall, this paper presents a practical solution toward more reliable VEC systems through the introduction of EmoCaliber. By verbalizing confidence alongside emotion predictions, the model is empowered to reflect the inherent subjectivity of emotion perception in images, effectively alleviating a long-standing challenge while remaining compatible with existing evaluation protocols. To support this goal, we develop a three-stage training framework and a complementary data construction pipeline that yields a high-quality VEC-CoT dataset. We further unify six widely used VEC datasets into VECBench, enabling fair and systematic evaluation of VEC in MLLMs. Extensive experiments demonstrate that EmoCaliber consistently delivers superior performances in both emotion prediction and confidence estimation, while also revealing insights into different optimization configurations. Through EmoCaliber and VECBench, we hope to establish solid baselines for future research and facilitate the development of reliable VEC systems.

\section*{Acknowledgment}
This work is supported by the National Natural Science Foundation of China (Grant No. 62376266 \& 62406318).

\bibliography{main}
\bibliographystyle{iclr2026_conference}

\end{document}